\documentclass[conference]{IEEEtran}
\usepackage{times}

\usepackage[numbers]{natbib}
\usepackage{multicol}
\usepackage[bookmarks=true]{hyperref}
\usepackage{graphicx}
\usepackage{amsmath}
\usepackage{textcomp}
\usepackage{color}
\usepackage[normalem]{ulem}

\begin{document}

\title{Refined Motion Compensation with Soft Laser Manipulators using Data-Driven Surrogate Models}




%
\author{\authorblockN{Yongjun Yan\authorrefmark{1}\authorrefmark{3},
Qingpeng Ding\authorrefmark{1},
Mingwu Li\authorrefmark{2}, 
Junyan Yan\authorrefmark{1} and
Shing Shin Cheng\authorrefmark{1}\authorrefmark{3}}
\authorblockA{\authorrefmark{1}Department of Mechanical and Automation Engineering\\
The Chinese University of Hong Kong,
Hong Kong SAR, China\\ Email: sscheng@cuhk.edu.hk}
\authorblockA{\authorrefmark{2}Department of Mechanics and Aerospace Engineering\\
Southern University of Science and Technology, 518055 Shenzhen, China}
\authorblockA{\authorrefmark{3}Multi-Scale Medical Robotics Center,
Hong Kong SAR, China}}

\maketitle

\begin{abstract}

Non-contact laser ablation, a precise thermal technique, simultaneously cuts and coagulates tissue without the insertion errors associated with rigid needles. Human organ motions, such as those in the liver, exhibit rhythmic components influenced by respiratory and cardiac cycles, making effective laser energy delivery to target lesions while compensating for tumor motion crucial. This research introduces a data-driven method to derive surrogate models of a soft manipulator. These low-dimensional models offer computational efficiency when integrated into the Model Predictive Control (MPC) framework, while still capturing the manipulator's dynamics with and without control input. Spectral Submanifolds (SSM) theory models the manipulator's autonomous dynamics, acknowledging its tendency to reach equilibrium when external forces are removed. Preliminary results show that the MPC controller using the surrogate model outperforms two other models within the same MPC framework. The data-driven MPC controller also supports a design-agnostic feature, allowing the interchangeability of different soft manipulators within the laser ablation surgery robot system.
\end{abstract}

\IEEEpeerreviewmaketitle

\section{Introduction}
Laser thermal ablation (LTA) and soft laser manipulators offer enhanced precision and safety during non-contact procedures~\cite{runciman2019soft,kundrat2019toward}, eliminating
ablation errors resulting from tissue deformation and needle deflection during insertion in traditional thermal ablation techniques with a rigid needle~\cite{crocetti2010quality}. However, precisely tracking tumor motion with a soft laser manipulator is challenging due to the infinite
degrees of freedom (DoFs) in the manipulator's structure and the consistently deformable nature~\cite{runciman2019soft}. Current analytical models and neural network approaches have limitations~\cite{yasa2023overview}, requiring models that balance accuracy and efficiency for effective motion compensation during laser ablation.

Data-driven modeling with structural priors provides explainable and parameterizable models for continuum robots, making them suitable for integration into mature control theories such as Model Predictive Control (MPC) to ensure robustness, stability, and invariance-based safety properties of the soft laser manipulator control system~\cite{hewing2020learning,markovsky2023data}. In soft robotics, Koopman operator theory has been used to develop linear and nonlinear models for MPC~\cite{bruder2020data}. While linear Koopman models offer an acceptable computational footprint but lack accuracy, nonlinear Koopman models are more accurate but computationally intensive for real-time implementation. Contrary to the Koopman theory, which lifts nonlinear system dynamics into a high-dimensional latent state space, recent studies have shown that the dynamics of high-dimensional systems can be learned on low-dimensional, attracting invariant manifolds called Spectral Submanifolds (SSM)~\cite{cenedese2022data,li2022nonlinear1}. Preliminary research indicates that integrating the SSM model reduction method into an optimal control framework can result in more accurate tracking performance, as demonstrated with a toy platform~\cite{alora2023practical}.

In this work, we introduce an MPC-based, data-driven control framework for motion compensation using a soft laser manipulator. By applying SSM theory to a soft robot in a medical context, we develop a low-dimensional dynamic model from vibration trajectories with and without cable tension to serve as the surrogate model of the soft laser manipulator (Fig.\ref{fig1} (a)). A tailored solver utilizing the sequential quadratic programming (SQP) algorithm addresses the optimal control problem with the learned surrogate dynamic model and constraints\cite{yan2023eco}. Suboptimal and warm-start features of the SQP algorithm are applied to further reduce the computational burden. Experiments demonstrate that our proposed controller framework with surrogate models achieves superior tracking accuracy in periodic trajectories mimicking tumor motion induced by respiration. Additionally, we validate the design-agnostic feature of the proposed controller in minimally invasive tumor ablation surgeries.

\section{Mechanisms of Soft Laser Manipulator}


\begin{figure*}
	\begin{center}
		\includegraphics[width=2\columnwidth]{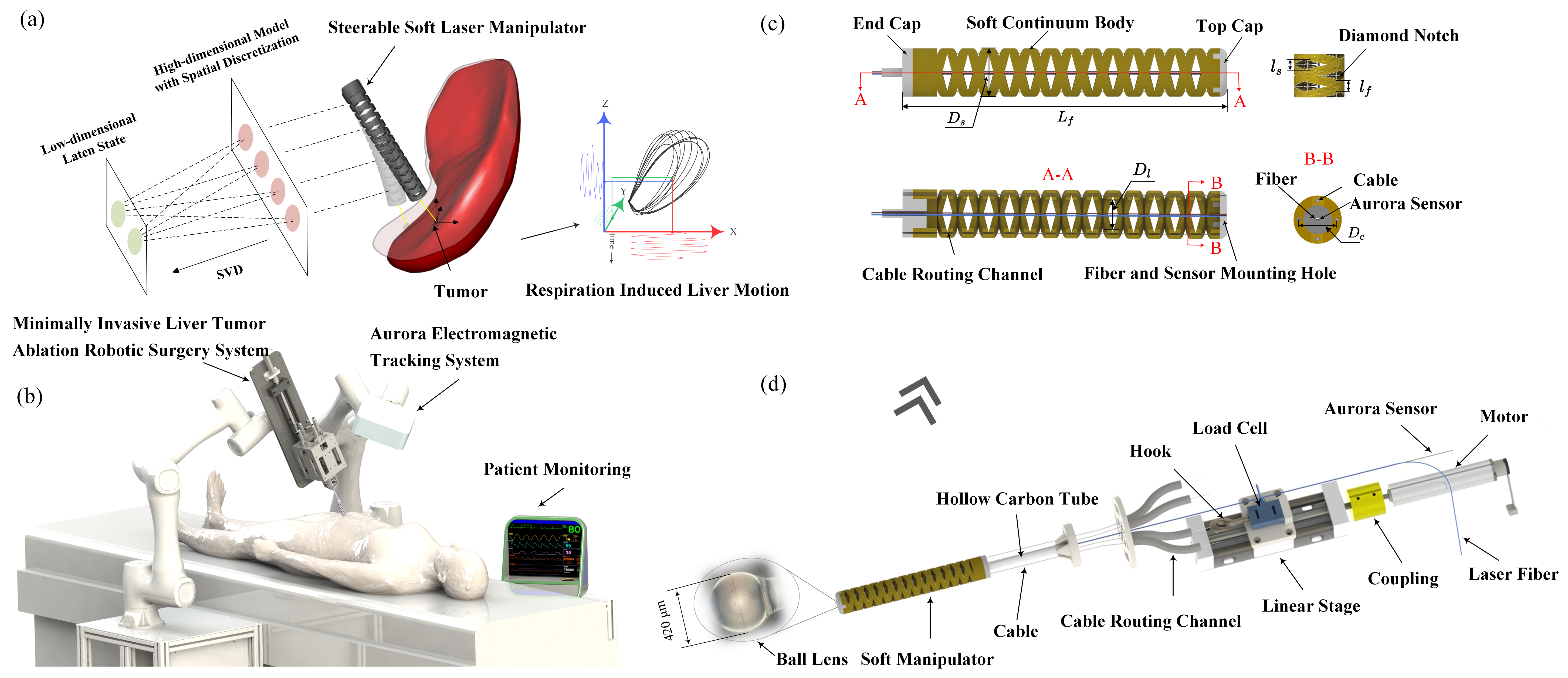}\vspace{-9pt}
		\caption{Schematic diagram of a soft laser manipulator in minimally invasive liver tumor ablation surgery. (a) Demonstration of the motion compensation mechanism with the surrogate models of the soft laser manipulator. (b) Clinical setup of the minimally invasive tumor ablation surgery with soft laser manipulator. (c) CAD models of the soft laser manipulator robotics system. (d) Exposed electro-mechanical subunits of one of the cable actuation systems. The four cable actuation systems are mounted in pairwise opposition in the cable actuation modules.}
	     \label{fig1} \vspace{-10pt}
	\end{center}
\end{figure*}

The clinical setup of the minimally invasive tumor ablation surgery using the soft laser manipulator is shown in Fig.\ref{fig1} (b). The surgical robot equipped with the laser fiber is inserted through a small incision to deliver non-contact focused laser ablation to a peripheral liver tumor site without penetrating the liver tissue~\cite{fusaglia2016clinically}.

The cable-driven soft laser manipulator is designed with a $D_s=14.0 ~\text {mm}$ outer diameter, a $D_l=8.4 ~\text {mm}$ lumen diameter, and a $L_f=94.0 ~\text {mm}$ bending length. The end effector, shown in Fig.\ref{fig1}(c), consists of a soft continuum body, a top cap, and an end cap. The manipulator is made from materials with significant stiffness differences: the soft body is vacuum-molded from polyurethane elastomer with a hardness of 70 Shore A, and the top and end caps are 3D-printed from photosensitive resin. To enable two degrees of freedom, four 1.2 mm diameter holes are spaced $90^{\circ}$ apart to create channels for the driving cables. Thirteen diamond-shaped notches (1.2 mm length, 11.0 mm pitch) along each channel provide anisotropic stiffness and directional deformation. The actuation setup for one cable is shown in Fig.\ref{fig1}(d). A DC motor-driven lead screw linear stage pulls and releases the cable along the manipulator's length, with a load cell measuring tension at the cable’s proximal end. A 1.0 mm diameter hole at the center of the top cap mounts the laser fiber and Aurora sensor. A 420 µm diameter ball lens focuses the laser light on the specimen surface. The end cap supports the continuum body under cable tension, allowing controlled movement.

The diamond-shaped notches along each cable channel are designed for controllable bending but also introduce anisotropic stiffness and damping properties, which are challenging to integrate into current dynamic control paradigms. Additionally, the slender, deformable structure of the soft laser manipulator results in twisting and shearing under transverse gravity loads and cable action, leading to kinematic shifts in tip orientation. Precise motion compensation in laser ablation therapy requires the manipulator's controller to manage a high-frequency, small-diameter 3D periodic trajectory, necessitating specialized control strategies.

\section{Controller Implementation Pipeline}

To illustrate the training process of data-driven surrogate models for the soft manipulator using low-dimensional SSM theory and to explain how SQP solves the optimal control problem, the implementation process of the MPC controller with the surrogate model is introduced in a pipeline style, as shown below.

In \textbf{Step 1}, snapshots are constructed by allowing the soft manipulator to decay from 18 different initial points to learn its autonomous dynamics. The Cartesian positions of two feature points, 1 mm apart along the longitudinal direction at the tip of the soft manipulator, are measured using an Aurora 6 DoF sensor. A time-delay of $d = 1$ is applied to the 18 decaying trajectories from these two feature points to derive the training snapshot.

In \textbf{Step 2}, Singular Value Decomposition (SVD) is performed, and the leading $n$ modes of the snapshot are selected to capture the majority of the decay pattern of the soft manipulator dynamics. Each trajectory $\mathbf{T}_k$ in the observable space is then projected onto the leading $n$ modes to obtain low-dimensional coordinates via the transformation: \(\mathbf{X}_k = \tilde{\mathbf{V}}^T_n \mathbf{T}_k\)~\cite{cenedese2022data}. The reduced dynamics on the manifold are learned via ridge regression,
\begin{equation}
\mathbf{R}^*=\underset{\mathbf{R}}{\arg \min } \sum_{k}\left\|\dot{\mathbf{T}}_k^d-\mathbf{R}_1 \mathbf{X}_k-\mathbf{R}_{2: n_r}\mathbf{X}_k^{2: n_r}\right\|_2^2,
\end{equation}
where all the eigenvalues of the linear state matrix $\mathbf{R}_1$ have negative real parts (these eigenvalues are also associated with the singular values), ensuring that the SSM attracts trajectories to the equilibrium points.

In \textbf{Step 3}, the control effect is incorporated into the dynamics equation by regressing the control matrix $\mathbf{C}_r$. A new dataset is collected by generating a smooth sequence of inputs through periodic actuation at various control amplitudes and recording the corresponding observed state trajectories. The mapping and updated dynamics equation with the control input is reorganized as:
\begin{equation}
\dot{\mathbf{x}}=\mathbf{r}(\mathbf{x},\mathbf{u})=\mathbf{R}_1 \mathbf{x}+\mathbf{R} \mathbf{x}^{2: n_r}+\mathbf{C}_{\mathbf{r}} \mathbf{u} .
\end{equation}
where \(\mathbf{x}\) and \(\mathbf{u}\) represent the state variables in the low-dimensional coordinates, and control variables, respectively. Considering the delay-embedding can not be performed with the current state measures, a mapping from the observable space to the low-dimensional space is fitted to facilitate the formulation of the optimal control problem:
\begin{equation}
\mathbf{x}=\mathbf{v}_z(\mathbf{z})=\tilde{\mathbf{V}}_0 \mathbf{z}+\tilde{\mathbf{V}} \mathbf{z}^{2: n_v}
\end{equation}
where \(\mathbf{z}\) represent the state variables in the observable space.

In \textbf{Step 4}, the synthesized data-driven surrogate model is integrated into an MPC framework to calculate optimal cable tension for motors, steering the laser dot to track tumor motion. MPC is chosen because it: 1) proactively adjusts cable tension by considering look-ahead trajectory information, compensating for lower-level tracking delays, and 2) naturally handles cable tension constraints. The optimal control problem is formulated as a nonlinear programming problem with a multiple-shooting method:
\begin{equation}
\min _{x(\cdot|j)u(\cdot|j)} \left\|\delta \mathbf{x}\left(N|j\right)\right\|_{\mathbf{Q}_f}^2 + \sum_{i=0}^{N-1}\|\delta \mathbf{x}(i|j)\|_{\mathbf{Q}}^2
\end{equation}
subject to
\begin{equation}
\begin{aligned}
& \mathbf{x}(0|j)=\mathbf{v}_z(\mathbf{z}(0|j)),\\
& {\mathbf{x}}(i+1|j)=  \mathbf{x}(i|j) +\mathbf{r}(\mathbf{x}(i|j),\mathbf{u}(i|j))\Delta t , i=0: N_{h}-1\\
& \mathbf{x}_{ref}(i|j)=\mathbf{v}_z(\mathbf{z}_{ref}(i|j)), i=0: N_{h}\\
& u(i|j) \in \mathcal{U}, i=0: N_{h}.
\end{aligned}
\end{equation}
Here, \(\delta \mathbf{x}(i|j) = \mathbf{x}(i|j) - \mathbf{x}_{ref}(i|j)\) represents the tracking error in low-dimensional coordinates. Unlike the approach in~\cite{alora2023practical}, which formulates the cost function in observable space, this research formulates the tracking error in low-dimensional space by mapping the reference trajectory to the low-dimensional space in advance to avoid the numerical instability of the SQP solver. A tailored solver based on the SQP algorithm is employed to solve the nonlinear programming problem, leveraging suboptimal and warm-start features to reduce computational burden. 

\section{Experimental Evaluation}


\begin{figure}[ht!]
	\begin{center}
		\includegraphics[width=1\columnwidth]{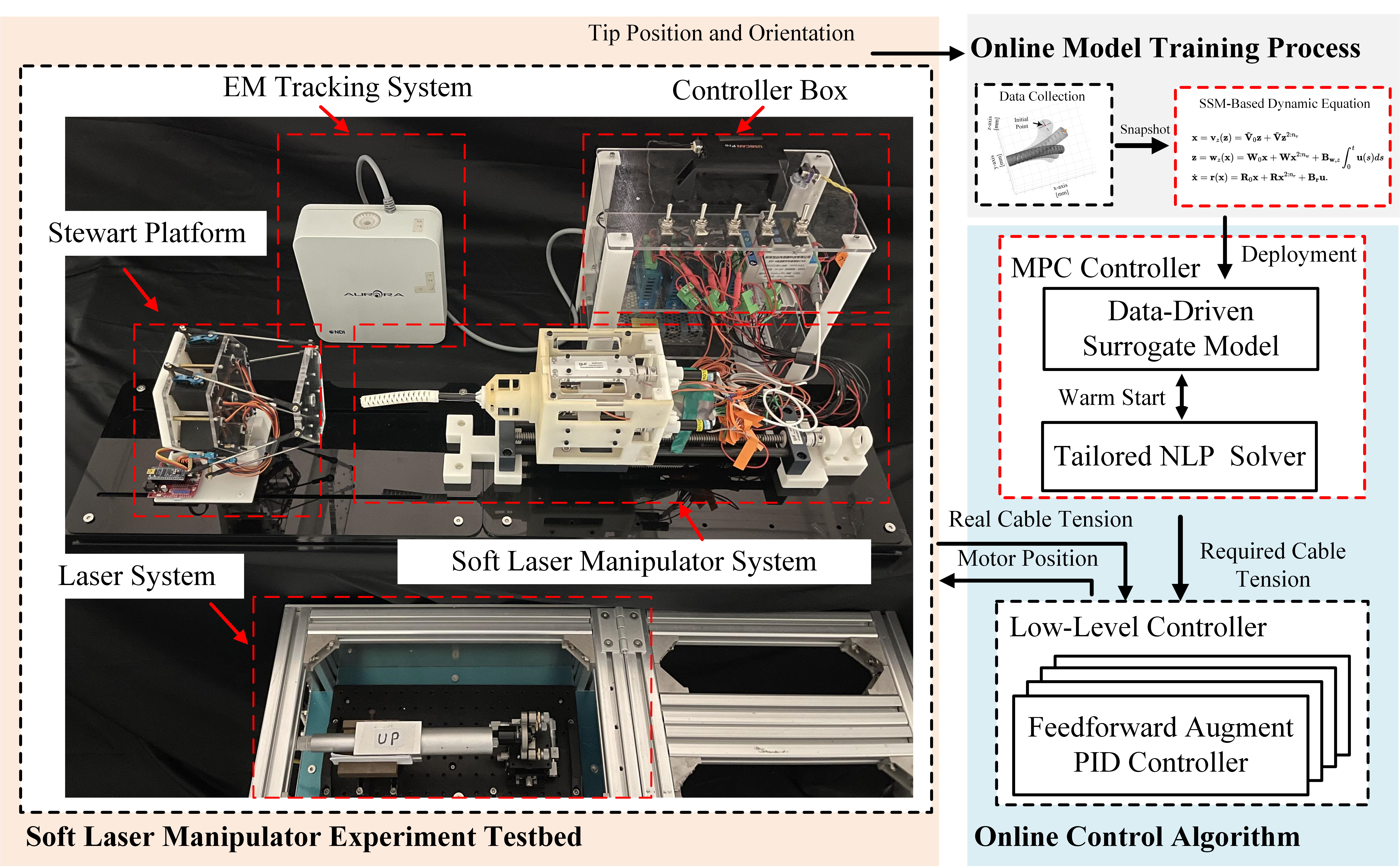}\vspace{-9pt}
		\caption{Experimental platform for the soft laser manipulator.}
	     \label{fig2} \vspace{-10pt}
	\end{center}
\end{figure}

A testbed (Fig.\ref{fig2}) was developed to evaluate the motion compensation performance of the MPC controller with the surrogate model for the soft laser manipulator. At the far end of the manipulator is a Stewart platform, replicating tumor motion induced by heartbeat and respiration.

\subsection{Control Performance Comparison with Other Models}

Two other MPC controllers, namely the Constant Curvature Model-based MPC (CC-MPC) controller~\cite{della2021model} and the Linear Koopman Model-based MPC (LK-MPC) controller~\cite{bruder2020data}, are benchmarked to assess the efficacy of integrating the low-dimensional surrogate model into the MPC scheme (SSM-MPC).


\begin{figure}[ht!]
	\begin{center}
		\includegraphics[width=1\columnwidth]{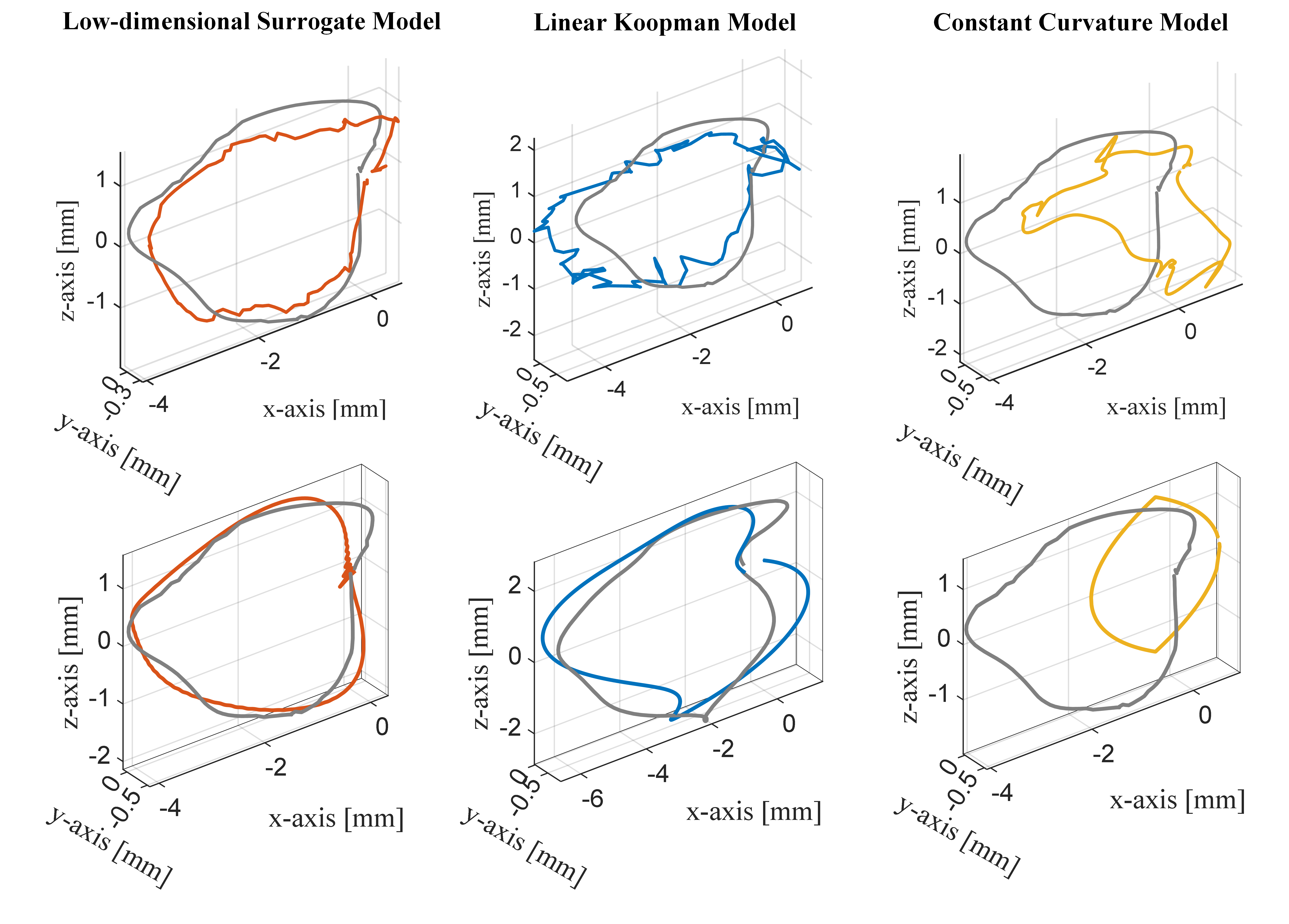}\vspace{-9pt}
		\caption{Reference tracking performance comparison of the SSM-MPC (row 1, column 1, red), LK-MPC (row 1, column 2, blue), and CC-MPC (row 1, column 3, yellow). Open-loop prediction accuracy comparison using identical initial states and control sequences (row 2).}
	     \label{fig3} \vspace{-10pt}
	\end{center}
\end{figure}

As shown in Fig.\ref{fig3}, the SSM-MPC controller with the low-dimensional surrogate model achieved the best tracking performance, with an overall average error of 0.27 mm, followed by the LK-MPC controller with an average error of 0.36 mm. 
The 25\% tracking error reduction of the SSM-MPC controller is significant, as it is achieved using a more accurate model prediction with only a three-dimensional SSM model compared to the 10-dimensional linear Koopman model with Fourier basis functions (row 2 in Fig.\ref{fig3}). This result aligns with the findings in~\cite{bruder2020data} that the linear Koopman model cannot capture all the nonlinearity of the soft manipulator. Therefore, the proposed SSM-MPC controller outperforms the LK-MPC in tracking accuracy while avoiding the computational burden of nonlinear Koopman models.

The CC-MPC controller exhibited the worst tracking performance with an average error of 0.58 mm, nearly twice that of the SSM-MPC controller (Fig.\ref{fig3}). This poor performance is due to the inaccuracy of the constant curvature model, where the constant curvature assumption fails in the experimental setup of the soft laser manipulator due to factors such as gravity and anisotropic properties.

\subsection{Design Agnostic Feature Evaluation}


\begin{figure}[ht!]
	\begin{center}
		\includegraphics[width=1\columnwidth]{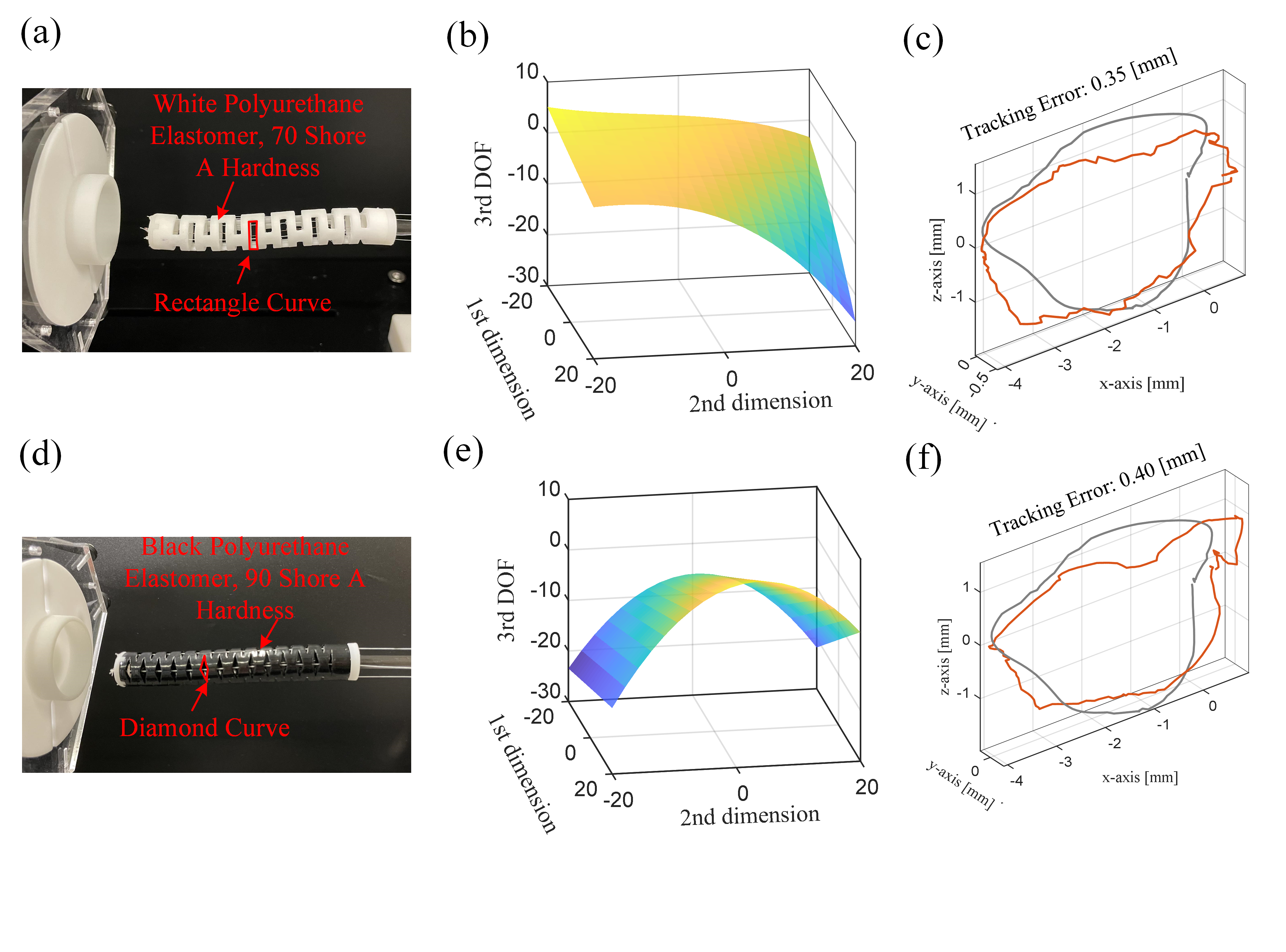}\vspace{-9pt}
		\caption{Design-agnostic feature evaluation of the SSM-MPC controller. (a) Soft manipulator with a rectangular curve and 70 Shore A hardness white polyurethane elastomer. (b) Learned submanifolds for the manipulator in (a). (c) Tracking performance of the SSM-MPC controller with the manipulator in (a). (d) Soft manipulator with a diamond curve and 90 Shore A hardness black polyurethane elastomer. (e) Learned submanifolds for the manipulator in (d). (f) Tracking performance of the SSM-MPC controller with the manipulator in (d).}
	     \label{fig4} \vspace{-10pt}
	\end{center}
\end{figure}

Compared with the first-principles models used in CC-MPC, the data-driven surrogate model that encodes the SSM as structural priors not only offers superior tracking performance for SSM-MPC but also maintains generalization capability across different soft manipulator designs. Similar to the parameter identification process for first-principles models, which involves parameters with physical interpretations, the surrogate model generalizes by learning different submanifolds with updated snapshots that reflect specific design variations. This design-agnostic feature allows the SSM-MPC controller to function effectively for various soft manipulator designs tailored to different ablation tasks without requiring tedious parameter tuning.

The design agnostic capability of the SSM-MPC controller is assessed and depicted in Fig.\ref{fig4}. With two distinct designs as depicted in Fig.\ref{fig4} (a) and (d), the SSM-MPC controller adeptly follows the trajectory, maintaining a tracking error of 0.40 mm. This performance is achieved by employing tailored SSM models (shown in Fig.\ref{fig4} (b) and (e)) corresponding to each specific design.

\section{Conclusion} 
\label{sec:conclusion}

The SSM is an ideal mathematical tool for depicting the attenuating oscillation pattern of the soft manipulator, as it effectively attracts trajectories to equilibrium points. The data-driven surrogate model leveraging SSM theory outperforms the linear Koopman and constant curvature models when integrated into the MPC framework for motion compensation of the soft manipulator. By learning the submanifolds of different soft manipulator designs from corresponding datasets, the data-driven MPC controller supports a design-agnostic feature, enabling the interchangeability of various soft manipulators within the laser ablation surgery robot system.

\section*{Acknowledgments}
Y. Yan acknowledges the support from the Multi-Scale Medical Robotics Center that made this work possible.

\bibliographystyle{plainnat}
\bibliography{references}

\begin{thebibliography}{13}
\providecommand{\natexlab}[1]{#1}
\providecommand{\url}[1]{\texttt{#1}}
\expandafter\ifx\csname urlstyle\endcsname\relax
  \providecommand{\doi}[1]{doi: #1}\else
  \providecommand{\doi}{doi: \begingroup \urlstyle{rm}\Url}\fi

\bibitem[Alora et~al.(2023)Alora, Cenedese, Schmerling, Haller, and Pavone]{alora2023practical}
John~Irvin Alora, Mattia Cenedese, Edward Schmerling, George Haller, and Marco Pavone.
\newblock Practical deployment of spectral submanifold reduction for optimal control of high-dimensional systems.
\newblock \emph{IFAC-PapersOnLine}, 56\penalty0 (2):\penalty0 4074--4081, 2023.

\bibitem[Bruder et~al.(2020)Bruder, Fu, Gillespie, Remy, and Vasudevan]{bruder2020data}
Daniel Bruder, Xun Fu, R~Brent Gillespie, C~David Remy, and Ram Vasudevan.
\newblock Data-driven control of soft robots using koopman operator theory.
\newblock \emph{IEEE Transactions on Robotics}, 37\penalty0 (3):\penalty0 948--961, 2020.

\bibitem[Cenedese et~al.(2022)Cenedese, Ax{\aa}s, B{\"a}uerlein, Avila, and Haller]{cenedese2022data}
Mattia Cenedese, Joar Ax{\aa}s, Bastian B{\"a}uerlein, Kerstin Avila, and George Haller.
\newblock Data-driven modeling and prediction of non-linearizable dynamics via spectral submanifolds.
\newblock \emph{Nature communications}, 13\penalty0 (1):\penalty0 872, 2022.

\bibitem[Crocetti et~al.(2010)Crocetti, de~Baere, and Lencioni]{crocetti2010quality}
Laura Crocetti, Thierry de~Baere, and Riccardo Lencioni.
\newblock Quality improvement guidelines for radiofrequency ablation of liver tumours.
\newblock \emph{Cardiovascular and interventional radiology}, 33:\penalty0 11--17, 2010.

\bibitem[Della~Santina et~al.(2023)Della~Santina, Duriez, and Rus]{della2021model}
Cosimo Della~Santina, Christian Duriez, and Daniela Rus.
\newblock Model-based control of soft robots: A survey of the state of the art and open challenges.
\newblock \emph{IEEE Control Systems Magazine}, 43\penalty0 (3):\penalty0 30--65, 2023.

\bibitem[Fusaglia et~al.(2016)Fusaglia, Hess, Schwalbe, Peterhans, Tinguely, Weber, and Lu]{fusaglia2016clinically}
Matteo Fusaglia, Hanspeter Hess, Marius Schwalbe, Matthias Peterhans, Pascale Tinguely, Stefan Weber, and Huanxiang Lu.
\newblock A clinically applicable laser-based image-guided system for laparoscopic liver procedures.
\newblock \emph{International journal of computer assisted radiology and surgery}, 11:\penalty0 1499--1513, 2016.

\bibitem[Hewing et~al.(2020)Hewing, Wabersich, Menner, and Zeilinger]{hewing2020learning}
Lukas Hewing, Kim~P Wabersich, Marcel Menner, and Melanie~N Zeilinger.
\newblock Learning-based model predictive control: Toward safe learning in control.
\newblock \emph{Annual Review of Control, Robotics, and Autonomous Systems}, 3:\penalty0 269--296, 2020.

\bibitem[Kundrat et~al.(2019)Kundrat, Schoob, Piskon, Gr{\"a}sslin, Schuler, Hoffmann, Kahrs, and Ortmaier]{kundrat2019toward}
Dennis Kundrat, Andreas Schoob, Thomas Piskon, Ren{\'e} Gr{\"a}sslin, Patrick~J Schuler, Thomas~K Hoffmann, L{\"u}der~A Kahrs, and Tobias Ortmaier.
\newblock Toward assistive technologies for focus adjustment in teleoperated robotic non-contact laser surgery.
\newblock \emph{IEEE Transactions on medical robotics and bionics}, 1\penalty0 (3):\penalty0 145--157, 2019.

\bibitem[Li et~al.(2022)Li, Jain, and Haller]{li2022nonlinear1}
Mingwu Li, Shobhit Jain, and George Haller.
\newblock Nonlinear analysis of forced mechanical systemswith internal resonance using spectral submanifolds, part i: Periodic response and forced response curve.
\newblock \emph{Nonlinear Dynamics}, 110\penalty0 (2):\penalty0 1005--1043, 2022.

\bibitem[Markovsky et~al.(2023)Markovsky, Huang, and D{\"o}rfler]{markovsky2023data}
Ivan Markovsky, Linbin Huang, and Florian D{\"o}rfler.
\newblock Data-driven control based on the behavioral approach: From theory to applications in power systems.
\newblock \emph{IEEE Control Systems Magazine}, 43\penalty0 (5):\penalty0 28--68, 2023.

\bibitem[Runciman et~al.(2019)Runciman, Darzi, and Mylonas]{runciman2019soft}
Mark Runciman, Ara Darzi, and George~P Mylonas.
\newblock Soft robotics in minimally invasive surgery.
\newblock \emph{Soft robotics}, 6\penalty0 (4):\penalty0 423--443, 2019.

\bibitem[Yan et~al.(2023)Yan, Li, Hong, Gao, Zhang, Chen, Sun, and Song]{yan2023eco}
Yongjun Yan, Nan Li, Jinlong Hong, Bingzhao Gao, Jia Zhang, Hong Chen, Jing Sun, and Ziyou Song.
\newblock Eco-coasting controller using road grade preview: Evaluation and online implementation based on mixed integer model predictive control.
\newblock \emph{IEEE Transactions on Vehicular Technology}, 2023.

\bibitem[Yasa et~al.(2023)Yasa, Toshimitsu, Michelis, Jones, Filippi, Buchner, and Katzschmann]{yasa2023overview}
Oncay Yasa, Yasunori Toshimitsu, Mike~Y Michelis, Lewis~S Jones, Miriam Filippi, Thomas Buchner, and Robert~K Katzschmann.
\newblock An overview of soft robotics.
\newblock \emph{Annual Review of Control, Robotics, and Autonomous Systems}, 6:\penalty0 1--29, 2023.

\end{thebibliography}

\end{document}